\documentclass[10pt,twocolumn,letterpaper]{article}

\usepackage{cvpr}
\usepackage{times}
\usepackage{epsfig}
\usepackage{graphicx}
\usepackage{amsmath}
\usepackage{amssymb}
\usepackage{appendix}
\usepackage{pbox}
\usepackage{mathabx}
\usepackage[pagebackref=true,breaklinks=true,letterpaper=true,colorlinks,bookmarks=false]{hyperref}

\cvprfinalcopy % *** Uncomment this line for the final submission

\def\cvprPaperID{****} % *** Enter the CVPR Paper ID here

\ifcvprfinal\fi
\begin{document}

%%%%%%%%% TITLE
\title{Semi-supervised learning based on generative adversarial network: a comparison between good GAN and bad GAN approach}

\author{Wenyuan Li\\
Electrical and Computer Engineering\\
University of California, Los Angeles\\
{\tt\small liwenyuan.zju@gmail.com}\thanks{W. Li and C. Arnold are the corresponding authors.}
\and
Zichen Wang\\
Bioengineering\\
University of California, Los Angeles\\
{\tt\small zcwang0702@g.ucla.edu} 
\and
Jiayun Li\\
Bioengineering\\
University of California, Los Angeles\\
{\tt\small jiayunli@ucla.edu}
\and
Jennifer Polson\\
Bioengineering\\
University of California, Los Angeles\\
{\tt\small jpolson@g.ucla.edu}
\and
William Speier\\
Radiology\\
University of California, Los Angeles\\
{\tt\small speier@ucla.edu}
\and
Corey Arnold\\
Radiology, Pathology, Bioengineering\\
University of California, Los Angeles\\
{\tt\small cwarnold@ucla.edu}
}

\maketitle
%\thispagestyle{empty}

%%%%%%%%% ABSTRACT
\begin{abstract}
Recently, semi-supervised learning methods based on generative adversarial networks (GANs) have received much attention. Among them, two distinct approaches have achieved competitive results on a variety of benchmark datasets. Bad GAN learns a classifier with unrealistic samples distributed on the complement of the support of the input data. Conversely, Triple GAN consists of a three-player game that tries to leverage good generated samples to boost classification results. In this paper, we perform a comprehensive comparison of these two approaches on different benchmark datasets. We demonstrate their different properties on image generation, and sensitivity to the amount of labeled data provided. By comprehensively comparing these two methods, we hope to shed light on the future of GAN-based semi-supervised learning. 
\footnote{This paper appears at CVPR 2019 Weakly Supervised Learning for Real-World Computer Vision Applications (LID) Workshop.}
\end{abstract}

%%%%%%%%% BODY TEXT
\section{Introduction}
Semi-supervised learning (SSL) aims to make use of large amounts of unlabeled data to boost model performance, typically when obtaining labeled data is expensive and time-consuming. Various semi-supervised learning methods have been proposed using deep learning and proven to be successful on several standard benchmarks. Weston \etal \cite{weston2012deep} employed a manifold embedding technique using the pre-constructed graph of unlabeled data; Rasmus \etal \cite{rasmus2015semi} used a specially designed auto-encoder to extract essential features for classification; Kingma and Welling \cite{kingma2013auto} developed a variational auto encoder in the context of semi-supervised learning by maximizing the variational lower bound of both labeled and unlabeled data; Miyato \etal \cite{miyato2018virtual} proposed virtual adversarial training (VAT) that tied to find a deep classifier, which had a good prediction accuracy on training data and meanwhile was less sensitive to data perturbation towards the adversarial direction.

Recently, generative adversarial networks (GANs) \cite{goodfellow2014generative}, have demonstrated their capability in SSL frameworks \cite{salimans2016improved, dai2017good, gan2017triangle, chongxuan2017triple, kumar2017semi, lecouat2018semi, li2018semi}. GANs are a powerful class of deep generative models that are able to model data distributions over natural images \cite{radford2015unsupervised, mirza2014conditional}. Salimans \etal first proposed to use GANs to solve a $(K + 1)$-class classification problem, where the dataset contained $K$ class originally and the additional $(K + 1)$th class consisted of the synthetic images generated by the GAN's generator. Later on, Li \etal \cite{chongxuan2017triple} realized that the generator and discriminator in \cite{salimans2016improved} may not be optimal at the same time (\textit{i.e.},  the discriminator was able to achieve good performance in SSL, while the generator may generate visually unrealistic images). They proposed a three-player game (Triple-GAN) to simultaneously achieve good classification results and obtained a good image generator. Dai \etal \cite{dai2017good} realized the same problem, but instead gave theoretical justifications of why using bad samples from the generator was able to boost SSL performance. Their model is called Bad GAN, which achieves state-of-the-art performance on multiple benchmark datasets. Another line of work focused on manifold regularization \cite{belkin2006manifold}. Kumar \etal \cite{kumar2017semi} estimated the manifold gradients at input data points and added an additional regularization term to a GAN, which promoted invariance of the discriminator to all directions in the data space. Lecouat \etal \cite{lecouat2018semi} performed manifold regularization by approximating the Laplacian norm that was easily computed within a GAN and achieved competitive results.

\begin{figure*}
\begin{center}
  \includegraphics[width=16cm,height=4cm]{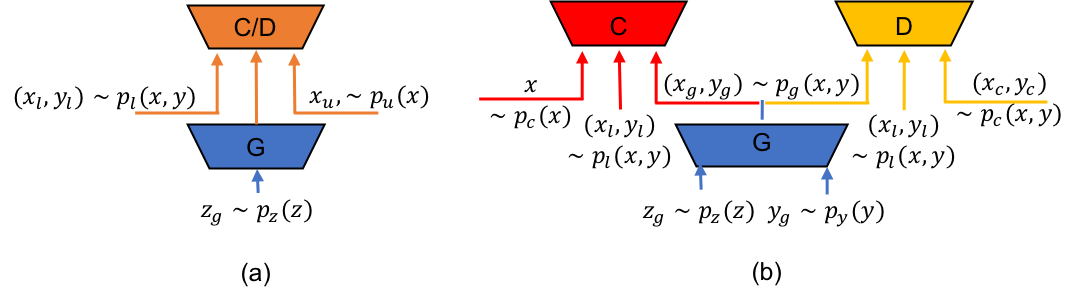}
  \caption{Network architecture of Bad GAN (a) and Good GAN (b). Bad GAN (a) consists of two parts: a generator \textit{G} aims to generates ``bad'' samples, and a discriminator/classifier \textit{D/C} that distinguishes real and fake samples and put the labeled samples into the right classes; Good GAN (b) consists of three parts: two conditional networks \textit{G} and \textit{C} that generate pseudo labels given real data and pseudo data given real labels respectively, and a separate discriminator \textit{D} that distinguish the generated data-label pair from the real data-label pair.} \label{Figure1}
\end{center}
\end{figure*}

In this paper, we focus on two GAN-based SSL models, Triple GAN and Bad GAN, and perform a comprehensive comparison between them. As both of models attempt to solve a similar issue in the original setting \cite{salimans2016improved} but are motivated by dissimilar perspectives, we believe that our comparison will provide insight for future SSL research. For simplicity, we refer to Triple GAN as Good GAN in contrast to Bad GAN. In Section \ref{Section2}, we briefly review the two models and their different approaches for solving loss function incompatibility; in Section \ref{Section3}, we show the network architecture we employed, benchmark datasets we used, and hyperparameters we selected in order to perform a fair comparison between these two models; in Section \ref{Section4}, we demonstrate our comparison results and discuss several important aspects we found for these two models; we conclude our paper in Section \ref{Section5}.  
%------------------------------------------------------------------------
\section{Related Work}\label{Section2}
\subsection{Bad GAN}
Suppose we have a classification problem that requires classifying a data point $\boldsymbol{x}$ into one of $K$ possible classes. A standard classifier takes in $\boldsymbol{x}$ as input and outputs a $K$-dimensional vector of logits $\{l_1, ..., l_K\}$. Salimens \etal \cite{salimans2016improved} extend the standard classifier \textit{C} by simply adding samples from the GAN generator \textit{G} to the dataset, labeling them as a new ``generated'' class $y = K + 1$, and correspondingly increasing the dimension of \textit{C}  output from $K$ to $K + 1$. The loss function $L_{C/D}$ for training \textit{C} (\textit{i.e.}, the extended discriminator \textit{D} from the GAN's perspective) then becomes

\begin{equation} \label{badgan_discriminator}
\begin{aligned}
L_{\textit{C/D}} &= L_{\text{supervised}} + L_{\text{unsupervised}}\\
L_{\text{supervised}} &= \mathop{\mathbb{E}}_{\boldsymbol{x}, y\sim p_{l}(\boldsymbol{x}, y)}[-\log (p_{C/D}(y|\boldsymbol{x}, y < K + 1))]\\
L_{\text{unsupervised}} &=\mathop{\mathbb{E}}_{\boldsymbol{x}\sim p_{u}(\boldsymbol{x})}[-\log (1 - p_{C/D}(y = K + 1|\boldsymbol{x}))]\\
 &+ \mathop{\mathbb{E}}_{\boldsymbol{x}\sim p_g(\boldsymbol{x})}[-\log (p_{C/D}(y = K + 1|\boldsymbol{x}))]
\end{aligned}
\end{equation}

The supervised loss term $L_\text{supervised}$ is a traditional cross-entropy loss that is applied to labeled data $(\boldsymbol{x}, y)\sim p_l(\boldsymbol{x}, y)$. The unsupervised loss requires \textit{C/D} to put the synthetic data from generator $\boldsymbol{x} \sim p_{g}(\boldsymbol{x})$ into the $(K+1)$th class, while putting the unlabeled data $\boldsymbol{x} \sim p_u(\boldsymbol{x})$ into the real $K$ classes. For the generator, \cite{salimans2016improved} found feature matching loss in Eq. \ref{feature-matching} is the best in practice, though they generated visually unrealistic images. The feature matching loss is,

\begin{equation} \label{feature-matching}
\begin{aligned}
L_{\textit{G}} &= \left\| \mathop{\mathbb{E}}_{\boldsymbol{x}\sim p_{u}}(\boldsymbol{f}(\boldsymbol{x})) - \mathop{\mathbb{E}}_{\boldsymbol{z}_g\sim p_{z}(z)}(\boldsymbol{f}(G(\boldsymbol{z}_g))) \right\|_2^2
\end{aligned}
\end{equation}

where $\boldsymbol{z}_g \sim p_z(\boldsymbol{z})$ is drawn from a simple distribution such as uniform.

On the basis of this formulation, Dai \etal \cite{dai2017good} give a theoretical justification on why the visually unrealistic images (\textit{i.e.}, ``bad'' samples) from the generator could help with SSL. Loosely speaking, the carefully generated ``bad'' samples along with the loss function design in Eq. \ref{badgan_discriminator} could force \textit{C}'s decision boundary to lie between the data manifolds of different classes, which in turn improves generalization of the classifier. Based on this analysis, they propose a Bad GAN model that learns a bad generator by explicitly adding a penalty term to generate ``bad'' samples. Their objective function of the generator becomes:

\begin{equation} \label{badgan_generator}
\begin{aligned}
L_{\textit{G}} &= -\mathcal{H}[p_{g}(\boldsymbol{x})] + \mathop{\mathbb{E}}_{\boldsymbol{x}\sim p_{g}(\boldsymbol{x})}(\log p^{pt}(x) \mathop{\mathbb{I}}[p^{pt}(x) > \epsilon]\\
&+ \left\| \mathop{\mathbb{E}}_{\boldsymbol{x}\sim p_{u}(\boldsymbol{x})}(\boldsymbol{f}(\boldsymbol{x})) - \mathop{\mathbb{E}}_{\boldsymbol{z}_g\sim p_{z}(z)}(\boldsymbol{f}(G(\boldsymbol{z}_g))) \right\|_2^2 \\
\end{aligned}
\end{equation}

where the first term measures the negative entropy of the generated samples and tries to avoid collapsing while increasing the coverage of the generator. The second term explicitly penalizes generated samples that are in high density areas by using a pre-trained model, and the third term is the same feature matching term as in Eq. \ref{feature-matching}.

\subsection{Good GAN}
Li \etal \cite{chongxuan2017triple} also noticed the same problem in \cite{salimans2016improved} as the generator and the discriminator have incompatible loss functions, but took a different approach to tackling this issue. Intuitively, assume the generator can generate good samples in the original settings of \cite{salimans2016improved}, the discriminator should identify these samples as fake samples as well as predict the correct class for the generated samples. To address the problem, \cite{chongxuan2017triple} present a three-player game called Triple-GAN that consists of a generator \textit{G}, a discriminator \textit{D}, and a separate classifier \textit{C}. \textit{C} and \textit{D} are two conditional networks that generate pseudo labels given real data and pseudo data given real labels respectively. To jointly evaluate the quality of the samples from the two conditional networks, a single discriminator \textit{D} is used to distinguish whether a data--label pair is from the real labeled dataset or not. We refer this model as Good GAN because one of the aims for this formulation is to obtain a good generator.

The authors prove that instead of competing equilibrium states as in \cite{salimans2016improved}, Good GAN has the unique global optimum for both \textit{C} and \textit{G}, \textit{i.e.}, $p(\boldsymbol{x}, y) = p_g(\boldsymbol{x}, y) = p_c(\boldsymbol{x}, y)$, the three joint distributions match one another. In other words, a good classifier will result in a good generator and vice versa. Furthermore, Good GAN is trained using the REINFORCE algorithm, in which it generates pseudo labels through \textit{C} for some unlabeled data and uses these pairs as positive samples to feed into \textit{D}. This is a key to the success of the model, as one of the crucial problems of SSL is the limited size of the labeled data. Figure \ref{Figure1} shows the network architecture of Good GAN and Bad GAN.

\section{Comparison Method}\label{Section3}
\subsection{Network Architecture}
In Bad GAN, the discriminator has two roles: to classify the real data into the right class and to distinguish the real samples from the fake samples.  For clarity, we refer to Bad GAN's discriminator as the classifier, since its input and output are exactly the same as the classifier in Good GAN due to the over-parameterization of the softmax layer \cite{salimans2016improved}.

To perform a fair comparison between Good GAN and Bad GAN, we use the same network architecture for the generator \textit{G} and the classifier \textit{C} in both models. We follow the architecture closely in \cite{chongxuan2017triple} to set up the additional discriminator \textit{D} in Good GAN. Both of them use Leaky-Relu activation and weight normalization to ease the difficulty of GAN's training. Implementing them using same architecture ideally avoids the possibility of using an architecture that is custom-tailored to work well with one or the other. Detailed model architectures can be found in the Appendix \ref{appA}. 

\subsection{Datasets}
Using the above-defined network architectures, we compare the two models on the widely adopted MNIST \cite{lecun1998gradient}, SVHN \cite{netzer2011reading}, and CIFAR10 \cite{krizhevsky2009learning} datasets. MNIST consists of 50,000 training samples, 10,000 validation samples, and 10,000 testing samples of handwritten digits of size $28 \times 28$. SVHN consists of 73,257 training samples and 26,032 testing samples. Each sample is a colored image of size $32 \times 32$, containing a sequence of digits with various backgrounds. CIFAR10 consists of colored images distributed across 10 general classes -- \textit{airplane}, \textit{automobile}, \textit{bird}, \textit{cat}, \textit{deer}, \textit{dog}, \textit{frog}, \textit{horse}, \textit{ship} and \textit{truck}. It contains 50,000 training samples and 10,000 testing samples of size $32 \times 32$. Following \cite{chongxuan2017triple}, we reserve 5,000 training samples from SVHN and CIFAR10 for validation if needed. For our CIFAR10 experiment, we perform zero-based component analysis (ZCA) \cite{laine2016temporal} as suggested in \cite{chongxuan2017triple} for the input of \textit{C}, but still generate and estimate the raw images using \textit{G} and \textit{D}.

We perform an extensive investigation by varying the amount of labeled data.  Following common practice, this is done by throwing away different amounts of the underlying labeled dataset \cite{salimans2016improved, pu2016variational, sajjadi2016mutual, tarvainen2017mean}. The labeled data used for training are randomly selected stratified samples unless otherwise specified. We perform our experiments on setups with 20, 50, 100, and 200 labeled examples in MNIST, 500, 1000, and 2000 labeled examples in SVHN, and 1000, 2000, 400, 8000 examples in CIFAR10. 

\subsection{Hyperparameter Selection}
For the hyperparameter selection such as learning rate and beta for Adam optimization, and the coefficient for each cost function term, we closely follow \cite{chongxuan2017triple, dai2017good}. In addition, we perform extensive study of the effects of batch size on performance for Bad GAN. As reported by \cite{lecouat2018semi}, Bad GAN training is sensitive to training batch size, and thus we vary batch size in the training phase and compare their final performances on MNIST and SVHN.

\section{Experimental Results and Discussion}\label{Section4}
\begin{table*}[!ht]
\caption{Test accuracy on semi-supervised MNIST. Results are averaged over 10 runs. * denotes the special selection of labeled data. See details in Section \ref{label_data_importance}.}
\label{Table1}
\centering
\begin{tabular}{ccccc}
\hline
Model & \multicolumn{4}{c}{\begin{tabular}[c]{@{}c@{}}Test accuracy for \\ a given number of labeled samples\end{tabular}} \\& 20& 50& 100& 200\\ \hline \hline
Bad GAN \cite{dai2017good}                &-                            &-                            & $\mathbf{99.21\pm0.01\%}$      &-                            \\
Triple GAN \cite{chongxuan2017triple}            & $95.19\pm4.95\%$                & $98.44\pm0.72\%$               & $99.09\pm0.58\%$               & $\mathbf{99.33\pm0.16\%}$      \\ \hline
Bad GAN (ours)         & $68.12\pm0.60\%$                     & $96.24\pm0.16\%$                    & $99.17\pm0.03\%$                    & $99.20\pm0.03\%$                    \\
Good GAN (ours)        & $\mathbf{95.93\pm4.45\%^{*}}$       & $\mathbf{98.68\pm1.12\%}$      & $99.07\pm0.46\%$               & $99.17\pm0.08\%$               \\ \hline
\end{tabular}
\end{table*}

\begin{table*}[!ht]
\caption{Test accuracy on semi-supervised SVHN. Results are averaged over 10 runs.}
\label{Table2}
\centering
\begin{tabular}{cccc}
\hline
Model & \multicolumn{3}{c}{\begin{tabular}[c]{@{}c@{}}Test accuracy for \\ a given number of labeled samples\end{tabular}} \\& 500 & 1000 & 2000                         \\ \hline \hline
Bad GAN\cite{dai2017good}                & -                            & $\mathbf{95.75 \pm 0.03\%}$             & -                   \\
Triple GAN\cite{chongxuan2017triple}             & -                            & $94.23 \pm 0.17\%$                     & -                            \\ \hline
Bad GAN (ours)         & $94.21 \pm 0.45\%$                      & $95.32 \pm 0.07 \%$                                              & $\mathbf{95.47 \pm 0.39\%}$             \\
Good GAN (ours)        & $\mathbf{94.67 \pm 0.12\%}$             & $95.30 \pm 0.38\%$                                              & $95.37 \pm 0.09\%$                      \\ \hline
\end{tabular}
\end{table*}

\begin{table*}[!ht]
\caption{Test accuracy on semi-supervised CIFAR10. Results are averaged over 10 runs.}
\label{Table3}
\centering
\begin{tabular}{ccccc}
\hline
Model & \multicolumn{4}{c}{\begin{tabular}[c]{@{}c@{}}Test accuracy for \\ a given number of labeled samples\end{tabular}} \\
                       & 1000                      & 2000                      & 4000                           & 8000                      \\ \hline \hline
Bad GAN \cite{dai2017good}                & -                         & -                         & $\mathbf{85.59\pm 0.03\%}$                   & -                         \\
Triple GAN \cite{chongxuan2017triple}             & -                         & -                         & $83.01 \pm 0.36\%$                   & -                         \\ \hline
Bad GAN (ours)         & $77.58 \pm 0.17\%$                   & $81.36 \pm 0.08\%$                   & $82.89 \pm 0.13\%$                        & $\mathbf{85.47 \pm 0.10\%}$ \\ \hline         Good GAN (ours)          &      $\mathbf{81.08 \pm 0.57\%}$              &   $\mathbf{81.79 \pm 0.37\%}$    & $82.82 \pm 0.41\%$                        &    $85.37 \pm 0.18\%$                       \\ \hline
\end{tabular}
\end{table*}

We implement Good GAN based on Tensorflow 1.10 \cite{girija2016tensorflow} and Bad GAN based on Pytorch 1.0 \cite{paszke2017automatic}. The generated images from $gG$ is not applied until the number of epochs reach a threshold that $gG$ could generate reliable image-lable pairs. We choose 200 in all three cases. All of the other hyperparameters including initial learning rate, maximum epoch number, relative weights and parameters in Adam \cite{kingma2014adam} are fixed according to \cite{salimans2016improved,chongxuan2017triple,dai2017good} across all of the experiments. 

\subsection{Classification}
We report our classification accuracy on the test set in Table \ref{Table1}, Table \ref{Table2} and Table \ref{Table3} for MNIST, SVNH and CIFAR10, respectively, along with the results reported in the original papers. The similarity of our results to those reported in the original papers suggests that our reproduced models are accurate instantiations of Good GAN and Bad GAN. Furthermore, we perform extensive study by varying the amount of labeled data and observe that Good GAN and Bad GAN behave quite differently under various circumstances.

First, with a medium amount of labeled data (\textit{e.g.}, MNIST with 100 or 200 labeled data, SVHN with more than 2000 labeled data, or CIFAR10 with more than 2000 labeled data), Bad GAN performs better than Good GAN. In fact, to the best of our knowledge, Bad GAN achieves the current state-of-the-art performance on those benchmark datasets. However, with low amounts of labeled data, Good GAN performs better, which demonstrates that Good GAN is less sensitive to the amount of labeled data than Bad GAN. One possible explanation is due to the use of the REINFORCE algorithm in Good GAN, because it generates pseudo labels through \textit{C} for some unlabeled data and use these pairs as positive samples of \textit{D}. Since \textit{C} converges quickly, this trick provides a clever way to enable the generator to explore a much larger data manifold that includes both the labeled and unlabeled data information. In other words, the classifier is able to provide pseudo labels for the unlabeled data, while the discriminator will judge if the pseudo labels are reliable or not throughout the training. This in return will affect the evolution of the generator, which will take advantage of the unlabeled data to generate good images. Generated good image-label pairs that implicitly contain unlabeled data information will eventually benefit the classifier. This works extremely well for relatively simple datasets like MNIST, as Good GAN is able to model the class-awarded data distribution through weak supervision. On the other hand, Bad GAN yields decreased performance when the amount of labeled data is low, as it does not have any mechanism to augment the information that could be used to train the classifier in this case.

\subsection{Generated Images}
\begin{figure*}
\begin{center}
  \includegraphics[width=14cm]{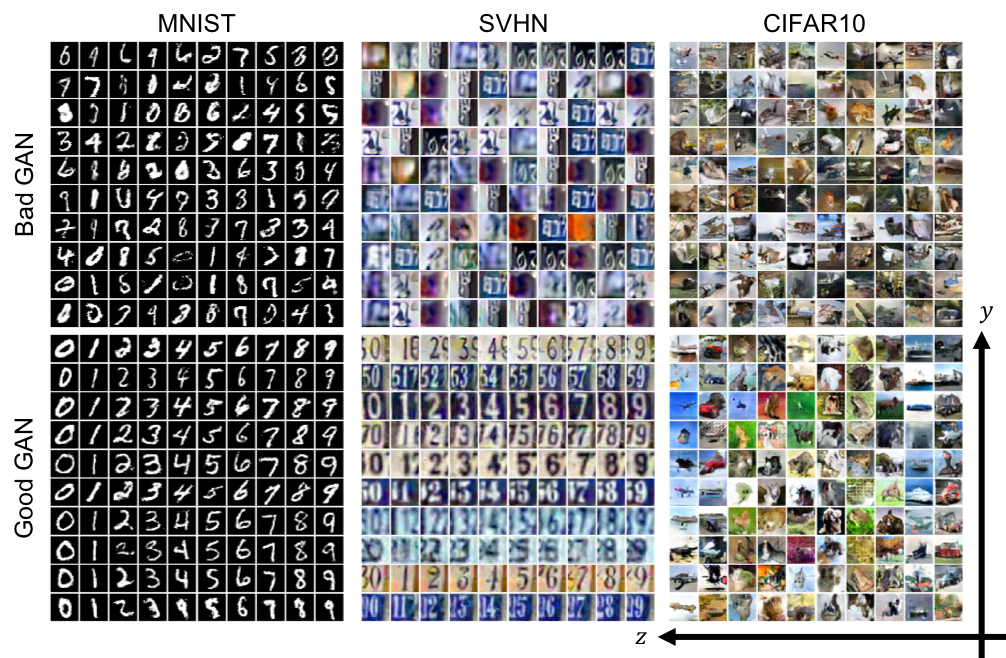}
  \caption{Generated images from both Bad GAN (top) and Good GAN (bottom). The images generated from Good GAN are produced by varying the class label $y$ in the vertical axis and the latent vector $z$ in the horizontal axis.} \label{Figure2}
\end{center}
\end{figure*}

\begin{figure*}
\begin{center}
  \includegraphics[width=13cm]{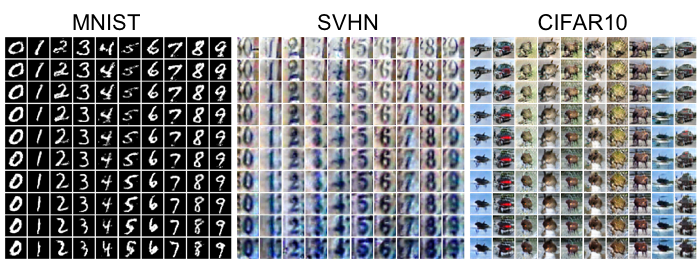}
  \caption{Class-conditional latent space interpolation. We first sample two random latent vectors $z$ and linearly interpolate them. Then we map these vectors to the image space conditioned on each class $y$. The vertical axis is the direction for latent vector interpolation while the horizontal axis is the direction for varying the class labels.} \label{Figure3}
\end{center}
\end{figure*}

In Figure \ref{Figure2}, we compare the quality of images generated by Good GAN and Bad GAN. As can be seen, Good GAN is able to generate clear images and meaningful samples conditioned on class labels, while Bad GAN generates ``bad'' images that look like a fusion of samples from different classes. In addition, Good GAN is able to disentangle classes and styles. In Figure \ref{Figure2} bottom, we vary the class label $y$ in the vertical axis and the latent vectors $z$ in the horizontal axis to generate the images. As shown in the figure, the latent vector $z$ encodes meaningful physical appearances, such as scale, intensity, orientation, color and so on, while the label $y$ controls the semantics of the generated images. Furthermore, Good-GAN can transition smoothly from one style to another with different visual factors without losing the label information as shown in Figure \ref{Figure3}. This proves that Good GAN can learn meaningful latent space representations instead of simply memorizing the training data.

\subsection{Importance of Selection of Labeled Data} \label{label_data_importance}
\begin{figure}
\begin{center}
  \includegraphics[width=8cm]{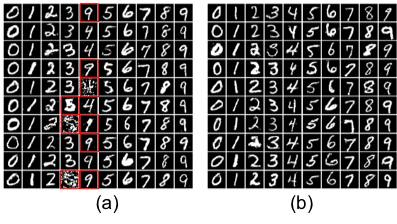}
  \caption{Two-runs of Good GAN model on MNIST dataset. (a) A single run where we randomly select 20 labeled data. The generator generates a lot of wrong images conditioned on the label and the classifier has lower performance. (b) Another run where we manually select 20 representative labeled examples. This time the generator is able to generate correct images, and the classifier achieves good classification performance.} \label{Figure4}
\end{center}
\end{figure}

Another interesting observation is that the selection of labeled data plays a crucial role for training Good GAN model in the low labeled data scenario. As mentioned above, the labeled data used for the training are randomly selected stratified samples, except for the MNIST-20 case. In this case, we found selecting representative labeled data to train is the key to achieving good performance. The reported accuracy in Table \ref{Table1} is averaged over 10 runs where we manually selected different representative labeled data in a stratified way. Figure \ref{Figure4} (a) shows a single run that uses randomly selected labeled data and does not achieve good results, while Figure \ref{Figure4} (b) shows another run that is able to achieve higher accuracy. The failure of the first run is due to the initial selections for digit 4 being similar to 9, causing the generator to generate many 9s when conditioned on label 4. The generator also generates low-quality images. We also report that with a random selection of 20 labeled data, the Good GAN was able to achieve $76.78 \pm 6.47 \%$ accuracy over 3 runs.

\subsection{Importance of Batch Size}
We found that batch size largely affect the final training results, in both Good GAN and Bad GAN. To investigate the effect of batch size on Bad GAN performance, we performed experiments with different batch size on MNIST (with 100 labeled samples) and SVHN (with 1000 labeled samples) using Bad GAN. As shown in Table \ref{Table4}, we empirically show that the performance of Bad GAN is sensitive to training batch size, and the optimal performance for each dataset is achieved with a batch size of 100.  

\begin{figure}
\begin{center}
  \includegraphics[width=8cm]{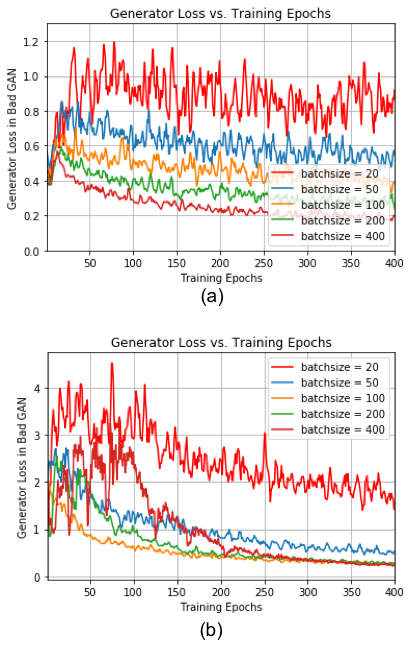}
  \caption{Batch size effect on generator loss in Bad GAN. The experiments are performed on (a) MNIST using 100 labeled samples and (b) SVHN using 1000 labeled samples.} \label{Figure5}
\end{center}
\end{figure}

To further understand the effect of the batch size on Bad GAN training, we present the generator loss with different batch sizes for MNIST and SVHN in Figure \ref{Figure5}.  The results indicate that smaller batch sizes lead to larger generator loss in the final stage of training. As  that generator loss mainly depends on the first-order feature matching loss in Bad GAN, an intuitive explanation could be that larger batch sizes reduce the variance of the sample mean, allowing the generator to quickly approximate the entire training set. This leads to smaller generator loss, especially when model training becomes more stable in the final stage.

As noted by \cite{dai2017good}, feature matching is performing distribution matching in a weak manner, which could be largely affected by batch size.  On one extreme, when the batch size is too small, the power of the generator in distribution matching is weak due to the excessive generator loss. Generated samples are therefore more likely to diverge from the manifold. Especially when data complexity increases, it is more difficult to minimize the KL divergence between the generator distribution and a desired complement distribution in Bad GAN, which could be one possible reason why model degradation is more significant on SVHN when using 20 batch size. On the other extreme, larger batch size leads to smaller generator loss, which comes with reduced diversity of generated samples. When the batch size is too large, the small generator loss will lead to a collapsed generator which fails to generate diverse samples that cover complement manifolds. As a result, the decision boundary between such missing manifolds becomes under-determined, which will also degrades model performance. We plot Bad GAN performance under different batch sizes for MNIST and SVHN in Appendix \ref{appB}. 

Based on our experience, Good GAN is best when we use a large batch size. Intuitively, a small batch size is not good for the REINFORCE algorithm adopted in Good GAN because a single wrong prediction of the unlabeled data will have a big impact on the weight update in each iteration.  We perform Good GAN experiments on SVHN using different batch size. The results are shown in Table \ref{Table5}. Empirically, we find that with small batch size, Good GAN is not able to generate good image-label pairs, hence the generated image-label pairs even hurt the classifier's performance when we use them to train. (See more details in Appendix \ref{appB}).

\begin{table*}[!ht]
\caption{Bad GAN performance versus batch size on MNIST and SVHN. The results are achieved using 100 labeled samples in MNIST and 1000 labeled samples in SVHN.}
\label{Table4}
\centering
\begin{tabular}{cccccc}
\hline
Batch size & 20 & 50 & 100 & 200 & 400    \\ \hline \hline
MNIST-100 &  $98.90 \pm 0.04\%$ &  $99.10 \pm 0.03\%$ &  $\mathbf{99.17 \pm 0.03\%}$ &  $99.16 \pm 0.03\%$  & $98.89 \pm 0.02\%$ \\ \hline
SVHN-1000 &  $93.35 \pm 0.05\%$ &  $95.29 \pm 0.03\%$ &  $\mathbf{95.56 \pm 0.02\%}$ &  $95.19 \pm 0.02\%$  &  $94.20 \pm 0.04\%$  \\ \hline
\end{tabular}
\end{table*}

\begin{table}[!ht]
\caption{Good GAN performance versus batch size on SVHN. The results are achieved using 1000 labeled samples in SVHN.}
\label{Table5}
\centering
\begin{tabular}{cccc}
\hline
Batch size & 20 & 50 & 100   \\ \hline \hline
SVHN-1000 &  $92.47\%$ &  $92.59\%$ &  $\mathbf{95.30\%}$  \\ \hline
\end{tabular}
\end{table}
\section{Conclusion}\label{Section5}
In this paper, we systematically and extensively compared two GAN-based SSL methods, Good GAN and Bad GAN, by applying these two models with commonly-used benchmark datasets. We illustrate the distinct characteristics of the images they generated, as well as each model’s sensitivity to varying the amount of labeled data used for training. In the case of low amounts of labeled data, model performance is contingent on the selection of labeled samples; that is, selecting non-representative samples results in generating incorrect image-label pairs and deteriorating classification performance. Furthermore, selecting the optimal batch size is crucial to achieve good results in both models. Notably, Good GAN and Bad GAN models can be used for complementary purposes; Good GAN generates good image-label pairs to train the classifier, while Bad GAN generates samples that force the decision boundary between data manifold of different classes. We envision that combining these two methods should yield further performance improvement in SSL.
\section*{Acknowledgements}
The authors would like to acknowledge support from the UCLA Radiology Department Exploratory Research Grant Program (16-0003) and NIH/NCI R21CA220352. This research was also enabled in part by GPUs donated by NVIDIA Corporation.
{\small
\bibliographystyle{ieee_fullname}
\bibliography{egbib}

\begin{thebibliography}{10}\itemsep=-1pt

\bibitem{belkin2006manifold}
Mikhail Belkin, Partha Niyogi, and Vikas Sindhwani.
\newblock Manifold regularization: A geometric framework for learning from
  labeled and unlabeled examples.
\newblock {\em Journal of machine learning research}, 7(Nov):2399--2434, 2006.

\bibitem{chongxuan2017triple}
LI Chongxuan, Taufik Xu, Jun Zhu, and Bo Zhang.
\newblock Triple generative adversarial nets.
\newblock In {\em Advances in neural information processing systems}, pages
  4088--4098, 2017.

\bibitem{dai2017good}
Zihang Dai, Zhilin Yang, Fan Yang, William~W Cohen, and Ruslan~R Salakhutdinov.
\newblock Good semi-supervised learning that requires a bad gan.
\newblock In {\em Advances in neural information processing systems}, pages
  6510--6520, 2017.

\bibitem{gan2017triangle}
Zhe Gan, Liqun Chen, Weiyao Wang, Yuchen Pu, Yizhe Zhang, Hao Liu, Chunyuan Li,
  and Lawrence Carin.
\newblock Triangle generative adversarial networks.
\newblock In {\em Advances in Neural Information Processing Systems}, pages
  5247--5256, 2017.

\bibitem{girija2016tensorflow}
Sanjay~Surendranath Girija.
\newblock Tensorflow: Large-scale machine learning on heterogeneous distributed
  systems.
\newblock {\em Software available from tensorflow. org}, 2016.

\bibitem{goodfellow2014generative}
Ian Goodfellow, Jean Pouget-Abadie, Mehdi Mirza, Bing Xu, David Warde-Farley,
  Sherjil Ozair, Aaron Courville, and Yoshua Bengio.
\newblock Generative adversarial nets.
\newblock In {\em Advances in neural information processing systems}, pages
  2672--2680, 2014.

\bibitem{kingma2014adam}
Diederik~P Kingma and Jimmy Ba.
\newblock Adam: A method for stochastic optimization.
\newblock {\em arXiv preprint arXiv:1412.6980}, 2014.

\bibitem{kingma2013auto}
Diederik~P Kingma and Max Welling.
\newblock Auto-encoding variational bayes.
\newblock {\em arXiv preprint arXiv:1312.6114}, 2013.

\bibitem{krizhevsky2009learning}
Alex Krizhevsky and Geoffrey Hinton.
\newblock Learning multiple layers of features from tiny images.
\newblock Technical report, Citeseer, 2009.

\bibitem{kumar2017semi}
Abhishek Kumar, Prasanna Sattigeri, and Tom Fletcher.
\newblock Semi-supervised learning with gans: Manifold invariance with improved
  inference.
\newblock In {\em Advances in Neural Information Processing Systems}, pages
  5534--5544, 2017.

\bibitem{laine2016temporal}
Samuli Laine and Timo Aila.
\newblock Temporal ensembling for semi-supervised learning.
\newblock {\em arXiv preprint arXiv:1610.02242}, 2016.

\bibitem{lecouat2018semi}
Bruno Lecouat, Chuan-Sheng Foo, Houssam Zenati, and Vijay~R Chandrasekhar.
\newblock Semi-supervised learning with gans: Revisiting manifold
  regularization.
\newblock {\em arXiv preprint arXiv:1805.08957}, 2018.

\bibitem{lecun1998gradient}
Yann LeCun, L{\'e}on Bottou, Yoshua Bengio, Patrick Haffner, et~al.
\newblock Gradient-based learning applied to document recognition.
\newblock {\em Proceedings of the IEEE}, 86(11):2278--2324, 1998.

\bibitem{li2018semi}
Wenyuan Li, Yunlong Wang, Yong Cai, Corey Arnold, Emily Zhao, and Yilian Yuan.
\newblock Semi-supervised rare disease detection using generative adversarial
  network.
\newblock {\em arXiv preprint arXiv:1812.00547}, 2018.

\bibitem{mirza2014conditional}
Mehdi Mirza and Simon Osindero.
\newblock Conditional generative adversarial nets.
\newblock {\em arXiv preprint arXiv:1411.1784}, 2014.

\bibitem{miyato2018virtual}
Takeru Miyato, Shin-ichi Maeda, Shin Ishii, and Masanori Koyama.
\newblock Virtual adversarial training: a regularization method for supervised
  and semi-supervised learning.
\newblock {\em IEEE transactions on pattern analysis and machine intelligence},
  2018.

\bibitem{netzer2011reading}
Yuval Netzer, Tao Wang, Adam Coates, Alessandro Bissacco, Bo Wu, and Andrew~Y
  Ng.
\newblock Reading digits in natural images with unsupervised feature learning.
\newblock In {\em Advances in neural information processing systems}, 2011.

\bibitem{paszke2017automatic}
Adam Paszke, Sam Gross, Soumith Chintala, Gregory Chanan, Edward Yang, Zachary
  DeVito, Zeming Lin, Alban Desmaison, Luca Antiga, and Adam Lerer.
\newblock Automatic differentiation in pytorch.
\newblock In {\em Advances in neural information processing systems Workshop},
  2017.

\bibitem{pu2016variational}
Yunchen Pu, Zhe Gan, Ricardo Henao, Xin Yuan, Chunyuan Li, Andrew Stevens, and
  Lawrence Carin.
\newblock Variational autoencoder for deep learning of images, labels and
  captions.
\newblock In {\em Advances in neural information processing systems}, pages
  2352--2360, 2016.

\bibitem{radford2015unsupervised}
Alec Radford, Luke Metz, and Soumith Chintala.
\newblock Unsupervised representation learning with deep convolutional
  generative adversarial networks.
\newblock {\em arXiv preprint arXiv:1511.06434}, 2015.

\bibitem{rasmus2015semi}
Antti Rasmus, Mathias Berglund, Mikko Honkala, Harri Valpola, and Tapani Raiko.
\newblock Semi-supervised learning with ladder networks.
\newblock In {\em Advances in neural information processing systems}, pages
  3546--3554, 2015.

\bibitem{sajjadi2016mutual}
Mehdi Sajjadi, Mehran Javanmardi, and Tolga Tasdizen.
\newblock Mutual exclusivity loss for semi-supervised deep learning.
\newblock In {\em 2016 IEEE International Conference on Image Processing
  (ICIP)}, pages 1908--1912. IEEE, 2016.

\bibitem{salimans2016improved}
Tim Salimans, Ian Goodfellow, Wojciech Zaremba, Vicki Cheung, Alec Radford, and
  Xi Chen.
\newblock Improved techniques for training gans.
\newblock In {\em Advances in neural information processing systems}, pages
  2234--2242, 2016.

\bibitem{tarvainen2017mean}
Antti Tarvainen and Harri Valpola.
\newblock Mean teachers are better role models: Weight-averaged consistency
  targets improve semi-supervised deep learning results.
\newblock In {\em Advances in neural information processing systems}, pages
  1195--1204, 2017.

\bibitem{weston2012deep}
Jason Weston, Fr{\'e}d{\'e}ric Ratle, Hossein Mobahi, and Ronan Collobert.
\newblock Deep learning via semi-supervised embedding.
\newblock In {\em Neural Networks: Tricks of the Trade}, pages 639--655.
  Springer, 2012.

\end{thebibliography}
}
\onecolumn
\appendixpage
\appendix
\section{Network Architecture}\label{appA}
We list the detailed architecture we used to compare Good GAN and Bad GAN on MNIST, SVHN, and CIFAR10 datasets in Table \ref{Table1app}, Table \ref{Table2app} and Table \ref{Table3app} respectively.

\begin{table*}[!ht]
\caption{MNIST}
\label{Table1app}
\centering
\begin{tabular}{c|c|c}
\hline
\textbf{Generator G} & \textbf{Classifier C} & \textbf{\begin{tabular}[c]{@{}c@{}}Discriminator D (Good GAN only)\end{tabular}}   \\ \hline
Input Label $y$, Noise $z$ & Input $28 \times 28$ Gray Image& Input $28 \times 28$ Gray Image, Label $y$\\ \hline
\begin{tabular}[c]{@{}c@{}}MLP 500 units, softplus, batch norm\\ \\ \\ MLP 500 units, softplus, batch norm\\ \\ \\ MLP 500 units, softplus, batch norm\end{tabular} & \begin{tabular}[c]{@{}c@{}}MLP 1000 units, lRelu, \\ Gaussian noise, weight norm\\ MLP 500 units, lRelu, \\ Gaussian noise, weight norm\\ MLP 250 units, lRelu, \\ Gaussian noise, weight norm\\ MLP 250 units, lRelu, \\ Gaussian noise, weight norm\\ MLP 250 units, lRelu, \\ Gaussian noise, weight norm\\ MLP 10 units, softmax, \\ Gaussian noise, weight norm\end{tabular} & \begin{tabular}[c]{@{}c@{}}MLP 1000 units, lRelu, \\ Gaussian noise, weight norm\\ MLP 500 units, lRelu, \\ Gaussian noise, weight norm\\ MLP 250 units, lRelu, \\ Gaussian noise, weight norm\\ MLP 250 units, lRelu, \\ Gaussian noise, weight norm\\ MLP 250 units, lRelu, \\ Gaussian noise, weight norm\\ MLP 12 units, sigmoid, \\ Gaussian noise, weight norm
\end{tabular}
\\
\hline
\end{tabular}
\end{table*}

\begin{table*}[!ht]
\caption{SVHN}
\label{Table2app}
\centering
\begin{tabular}{c|c|c}
\hline
\textbf{Generator G}& \textbf{Classifier C}& \textbf{Discriminator D (Good GAN only)}\\ \hline
Input Label $y$, Noise $z$ & Input $32 \times 32$ Colored Image & Input $32 \times 32$ Colored Image, Label $y$\\ \hline
\begin{tabular}[c]{@{}c@{}}MLP 8192 units, \\ Relu, batch norm\\ Reshape $512 \times 4 \times 4$\\ \\ \\ $5 \times 5$ deconv. 256. stride 2, \\ Relu, batch norm\end{tabular} & \begin{tabular}[c]{@{}c@{}}Gaussian noise, 0.2 dropout\\ $3 \times 3$ conv. 64. lRelu, weight norm\\$3 \times 3$ conv. 64. lRelu, weight norm\\$3 \times 3$ conv. 64. lRelu,  \\ stride 2, weight norm\\ 0.5 dropout\end{tabular}   & \begin{tabular}[c]{@{}c@{}}0.2 dropout\\ $3 \times 3$ conv. 32. lRelu, weight norm\\ $3 \times 3$ conv. 32. lRelu, \\stride 2, weight norm\\ 0.2 dropout\end{tabular}                 \\ \hline
\begin{tabular}[c]{@{}c@{}}$5 \times 5$ deconv. 128. stride 2, \\ Relu, batch norm\end{tabular}                                                               & \begin{tabular}[c]{@{}c@{}}$3 \times 3$ conv. 128. lRelu, weight norm\\ $3 \times 3$ conv. 128. lRelu, weight norm\\ $3 \times 3$ conv. 128. lRelu, \\stride 2, weight norm\\ 0.5 dropout\end{tabular}                               & \begin{tabular}[c]{@{}c@{}}$3 \times 3$ conv. 64. lRelu, weight norm\\ $3 \times 3$ conv. 64. lRelu, \\stride 2, weight norm\\ 0.2 dropout\end{tabular}                               \\ \hline
\begin{tabular}[c]{@{}c@{}}$5 \times 5$ deconv. 3. stride 2, \\ sigmoid, weight norm\end{tabular}                                                             & \begin{tabular}[c]{@{}c@{}}$3 \times 3$ conv. 128. lRelu, weight norm\\ $3 \times 3$ conv. 128. lRelu, weight norm\\ $3 \times 3$ conv. 128. lRelu, weight norm\\ \\ Global pool\\ MLP 10 units, softmax, weight norm\end{tabular} & \begin{tabular}[c]{@{}c@{}}$3 \times 3$ conv. 128. lRelu, weight norm\\ $3 \times 3$ conv. 128. lRelu, weight norm\\ \\ Global pool\\ MLP 1 unit, sigmoid, weight norm
\end{tabular}
\\
\hline
\end{tabular}
\end{table*}

\begin{table*}[!ht]
\caption{CIFAR10}
\label{Table3app}
\centering
\begin{tabular}{c|c|c}
\hline
\textbf{Generator G}& \textbf{Classifier C}& \textbf{Discriminator D (Good GAN only)}\\ \hline
Input Label $y$, Noise $z$ & Input $32 \times 32$ Colored Image & Input $32 \times 32$ Colored Image, Label $y$\\ \hline
\begin{tabular}[c]{@{}c@{}}MLP 8192 units, \\ Relu, batch norm\\ Reshape $512 \times 4 \times 4$\\ \\ \\ $5 \times 5$ deconv. 256. stride 2, \\ Relu, batch norm\end{tabular} & \begin{tabular}[c]{@{}c@{}}Gaussian noise, 0.2 dropout\\ $3 \times 3$ conv. 96. lRelu, weight norm\\$3 \times 3$ conv. 96. lRelu, weight norm\\$3 \times 3$ conv. 96. lRelu,  \\ stride 2, weight norm\\ 0.5 dropout\end{tabular}   & \begin{tabular}[c]{@{}c@{}}0.2 dropout\\ $3 \times 3$ conv. 32. lRelu, weight norm\\ $3 \times 3$ conv. 32. lRelu, \\stride 2, weight norm\\ 0.2 dropout\end{tabular}                 \\ \hline
\begin{tabular}[c]{@{}c@{}}$5 \times 5$ deconv. 128. stride 2, \\ Relu, batch norm\end{tabular}                                                               & \begin{tabular}[c]{@{}c@{}}$3 \times 3$ conv. 192. lRelu, weight norm\\ $3 \times 3$ conv. 192. lRelu, weight norm\\ $3 \times 3$ conv. 128. lRelu, \\stride 2, weight norm\\ 0.5 dropout\end{tabular}                               & \begin{tabular}[c]{@{}c@{}}$3 \times 3$ conv. 64. lRelu, weight norm\\ $3 \times 3$ conv. 64. lRelu, \\stride 2, weight norm\\ 0.2 dropout\end{tabular}                               \\ \hline
\begin{tabular}[c]{@{}c@{}}$5 \times 5$ deconv. 3. stride 2, \\ sigmoid, weight norm\end{tabular}                                                             & \begin{tabular}[c]{@{}c@{}}$3 \times 3$ conv. 192. lRelu, weight norm\\ $3 \times 3$ conv. 192. lRelu, weight ntheirorm\\ $3 \times 3$ conv. 192. lRelu, weight norm\\ \\ Global pool\\ MLP 10 units, softmax, weight norm\end{tabular} & \begin{tabular}[c]{@{}c@{}}$3 \times 3$ conv. 128. lRelu, weight norm\\ $3 \times 3$ conv. 128. lRelu, weight norm\\ \\ Global pool\\ MLP 1 unit, sigmoid, weight norm
\end{tabular}
\\
\hline
\end{tabular}
\end{table*}

\section{Batch Size Effect in Bad GAN}\label{appB}
\begin{figure}[!ht]
\begin{center}
  \includegraphics[width=\textwidth]{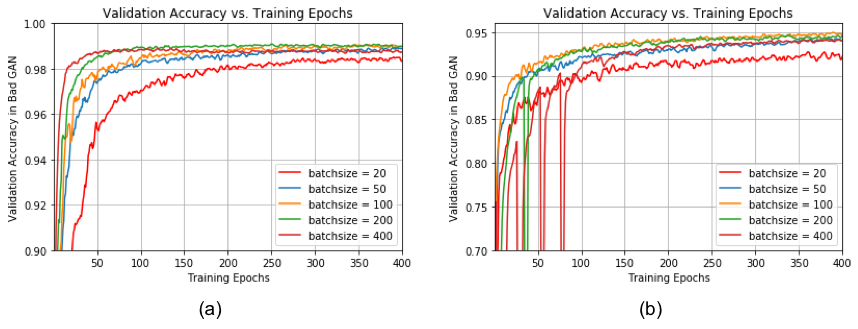}
  \caption{Batch size effect in Bad GAN. The classification accuracy over the initial 400 training epochs under different batch size. (a) The experiments are performed on MNIST dataset, using 100 labeled data. (b) The experiments are performed on SVHN dataset, using 1000 labeled data.} \label{Figure7}
\end{center}
\end{figure}

\begin{figure}
\begin{center}
  \includegraphics[width=13cm]{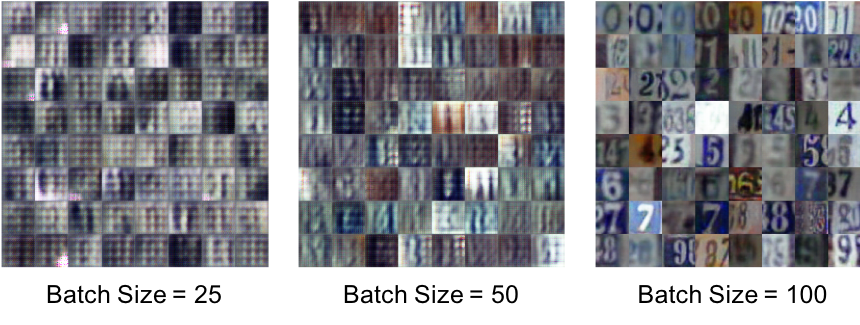}
  \caption{Batch size effect in Good GAN. With small batch size, Good GAN is not able to generate good image-label pairs. Experiments are performed on SVHN with $n = 1000$. All the images are generated at epoch $= 200$ when we start to use the generated image to train.} \label{Figure6}
\end{center}
\end{figure}

Figure \ref{Figure6} shows the classification accuracy under different batch size of Bad GAN during the first 400 epochs of training. As can be seen, the model performance is very sensitive to batch size. Figure \ref{Figure7} shows the generated images of Good GAN under different batch size. With small batch size, Good GAN is not able to generate good image-label pairs.

\end{document}